# Enhancing Fault-Tolerant Space Computing: Guidance Navigation and Control (GNC) and Landing Vision System (LVS) Implementations on Next-Gen Multi-Core Processors

Kyongsik Yun
Autonomous Systems
Jet Propulsion Laboratory
California Institute of Technology
Pasadena, CA, USA
kyongsik.yun@jpl.nasa.gov

David Bayard
Autonomous Systems
Jet Propulsion Laboratory
California Institute of Technology
Pasadena, CA, USA
david.s.bayard@jpl.nasa.gov

Gerik Kubiak
Autonomous Systems
Jet Propulsion Laboratory
California Institute of Technology
Pasadena, CA, USA
gerik.kubiak@jpl.nasa.gov

Austin Owens
Autonomous Systems
Jet Propulsion Laboratory
California Institute of Technology
Pasadena, CA, USA
austin.t.owens@jpl.nasa.gov

Andrew Johnson
Autonomous Systems
Jet Propulsion Laboratory
California Institute of Technology
Pasadena, CA, USA
aej@jpl.nasa.gov

Ryan Johnson
Autonomous Systems
Jet Propulsion Laboratory
California Institute of Technology
Pasadena, CA, USA
ryan.johnson@jpl.nasa.gov

Dan Scharf
Autonomous Systems
Jet Propulsion Laboratory
California Institute of Technology
Pasadena, CA, USA
daniel.p.scharf@jpl.nasa.gov

Thomas Lu
Autonomous Systems
Jet Propulsion Laboratory
California Institute of Technology
Pasadena, CA, USA
thomas.t.lu@jpl.nasa.gov

*Abstract*— **Future planetary exploration missions demand high-performance, fault-tolerant computing to enable autonomous Guidance, Navigation, and Control (GNC) and Lander Vision System (LVS) operations during Entry, Descent, and Landing (EDL). This paper evaluates the deployment of GNC and LVS algorithms on next-generation multi-core processors—HPSC, Snapdragon VOXL2, and AMD Xilinx Versal—demonstrating up to 15× speedup for LVS image processing and over 250× speedup for Guidance for Fuel-Optimal Large Divert (GFOLD) trajectory optimization compared to legacy spaceflight hardware. To ensure computational reliability, we present ARBITER (Asynchronous Redundant Behavior Inspection for Trusted Execution and Recovery), a Multi-Core Voting (MV) mechanism that performs real-time fault detection and correction across redundant cores. ARBITER is validated in both static optimization tasks (GFOLD) and dynamic closed-loop control (Attitude Control System). A fault injection study further identifies the gradient computation stage in GFOLD as the most sensitive to bit-level errors, motivating selective protection strategies and vector-based output arbitration. This work establishes a scalable and energy-efficient architecture for future missions, including Mars Sample Return, Enceladus Orbilander, and Ceres Sample Return, where onboard autonomy, low latency, and fault resilience are critical.**

## I. Introduction

Future planetary missions necessitate advanced, fault-tolerant computing platforms capable of real-time processing to support autonomous spacecraft operations. Critical components such as Guidance, Navigation, and Control (GNC), along with the Lander Vision System (LVS), are essential for successful Entry, Descent, and Landing (EDL) on planetary surfaces [1]–[3].

Traditional space computing platforms, including the RAD750 processor and Virtex-5 FPGA, are approaching obsolescence [4]. The RAD750, while historically reliable, offers limited processing capabilities that may not meet the demands of future missions requiring enhanced autonomy and real-time data processing [5]–[7]. Similarly, the Virtex-5 FPGA, though versatile, lacks the computational power and energy efficiency desired for next-generation applications [8].

Emerging computing platforms under evaluation include NASA's High-Performance Spaceflight Computing (HPSC) processor [7], [9], Qualcomm's Snapdragon VOXL2 [10], [11], and AMD's Xilinx Versal Adaptive SoCs [12], [13]. The HPSC is designed to deliver significant improvements in computational performance and fault tolerance, featuring a multi-core architecture tailored for space environments [14]. Snapdragon VOXL2 serves as a co-processor, integrating ARM cores, a Digital Signal Processor (DSP), and GPU capabilities to enhance onboard processing for applications such as computer vision and machine learning [15]. AMD's Xilinx Versal Adaptive SoCs combine adaptable processing with acceleration engines, offering a heterogeneous computing environment suitable for a wide range of space applications [16], [17]. A comparison of features and performance is shown in Table I.

TABLE I. COMPARISON OF LEGACY AND NEXT-GEN SPACE COMPUTING PLATFORMS

| Platform | Architecture & Compute Units | Performance Metrics | Fault Tolerance & Reliability |
|---|---|---|---|
| RAD750 (BAE Systems) | 1x PowerPC 750 @ 200 MHz; Scalar architecture; 250/150 nm process | 266-300 MIPS; ~0.25 GOPS; 5W power consumption; Single-threaded performance | Radiation hardened (2-10 kGy); EDAC on memory; Single-string CPU (no redundancy); Proven space heritage (20+ years) |
| HPSC (PIC64-HPSC) | 8x SiFive X280 RISC-V @ 1 GHz; 2x S7 system controllers; 512-bit vector extensions; 12nm FinFET process | 26,000 DMIPS (100x RAD750); ~25 GOPS scalar; 2 TOPS INT8 AI; 256 GFLOPS vector (theoretical) | Radiation hardened; ECC on all memories; Dual-core lockstep capable; Health monitoring & diagnostics; Dynamic power-performance scaling |
| Snapdragon (VOXL2) | 8x ARM cores (4+4); Kryo 585 @ 2.84 GHz; Adreno 650 GPU; Hexagon 698 DSP/NPU; 7nm process | ~100,000 MIPS (estimated); 15 TOPS INT8 AI; 1.0 TFLOP FP16 (GPU); ~0.6 TFLOP FP32 (GPU) | COTS (not rad-hard); Software watchdogs; Application redundancy; Requires external shielding; Thermal management |
| Xilinx Versal XQR (AMD) | 2x ARM Cortex-A72 @ 1.76 GHz; 2x ARM Cortex-R5F @ 750 MHz; Up to 400 AI Engines (AIE-ML); Programmable logic fabric; 7nm process | 133 TOPS INT8 AI; 3.2 TFLOPS FP32 (DSP); 13.6 TFLOPS INT8 (DSP); 31 TOPS INT8 (fabric); Adaptive performance scaling | Radiation tolerant (SEU/SEL hardened); ECC on all caches & memory; TMR implementation in fabric; Partial reconfiguration capability |

These next-generation processors provide enhanced computational power, fault tolerance, and efficiency, positioning them as strong candidates for future planetary missions [18], [19]. Their capabilities enable high-speed autonomous decision-making under stringent power and space constraints, addressing the increasing complexity and data processing requirements of upcoming exploration endeavors.

## II. GNC AND LVS IMPLEMENTATION ON NEXT-GEN PROCESSORS

The LVS image processing pipeline involves computationally expensive operations such as Fast Fourier Transform (FFT)-based correlation to match onboard images with pre-stored maps [20], [21]. The existing Vision Compute Element (VCE) on Mars 2020, which utilizes a RAD750 processor and a Virtex-5 FPGA, completes Forward FFT operations in 107 milliseconds and FFT correlation in 153 milliseconds per image. Implementing these algorithms on next-generation processors significantly reduces execution times, enabling real-time landmark correlation and hazard avoidance during planetary landing.

The Guidance for Fuel-Optimal Large Divert (GFOLD) algorithm, responsible for computing fuel-optimal guidance trajectories, also benefits from next-generation processors [22]–[25]. On a RAD750 processor, GFOLD takes approximately 7.52 seconds to compute trajectories with 2200 solution variables. Next-gen platforms reduce this computation to as little as 0.03 seconds, making in-flight adaptive trajectory optimization feasible even under real-time mission constraints. The performance gain for VOXL2 and VERSAL chips as compared to the original VCE processor is shown in Table II.

TABLE II. LVS PERFORMANCE GAINS ON NEXT-GEN PROCESSORS

| Function | VCE (RAD750 + Virtex-5) | VOXL2 | Speed up | Versal | Speed up |
|---|---|---|---|---|---|
| Forward FFT (1024x1024) | 107 ms | 14 ms | 7.6x | 6.79 ms | 15.8x |
| FFT Correlate (1024x1024) | 153 ms | 27 ms | 5.67x | 23.5 ms | 6.52x |
| GFOLD (400 vars) | 800 ms | 3 ms | 266x | 15 ms | 53x |
| GFOLD (2200 vars) | 7520 ms | 30 ms | 250x | 150 ms | 50x |

The performance evaluation of next-generation processors for onboard guidance and control demonstrates significant advancements in computational efficiency, achieving the required throughput for real-time operations. Figure 1 presents the log mean steady-state runtime as a function of solution variable size across multiple processors, including RAD750, BeagleV, VOXL 2, Versal, Snapdragon, and desktop-class processors. The results indicate that legacy spaceflight hardware, such as RAD750, exhibits execution times that are marginally feasible for Mars Science Laboratory (MSL) and Mars 2020 (M2020) missions but exceed the desired threshold for future deep space applications. In contrast, modern processors such as Snapdragon and Versal demonstrate execution times well within the mission goals of 0.25 seconds, with several configurations approaching the stretch goal of 0.03 seconds. These improvements enable real-time trajectory optimization for extended-range autonomous guidance, supporting the operational requirements of next-generation planetary landers and rovers. Specifically, for mission scenarios requiring larger solution variable sizes, such as Europa and Enceladus landers (~30 km scales), the performance of these processors ensures feasibility for computationally intensive onboard guidance algorithms like the G-FOLD framework.

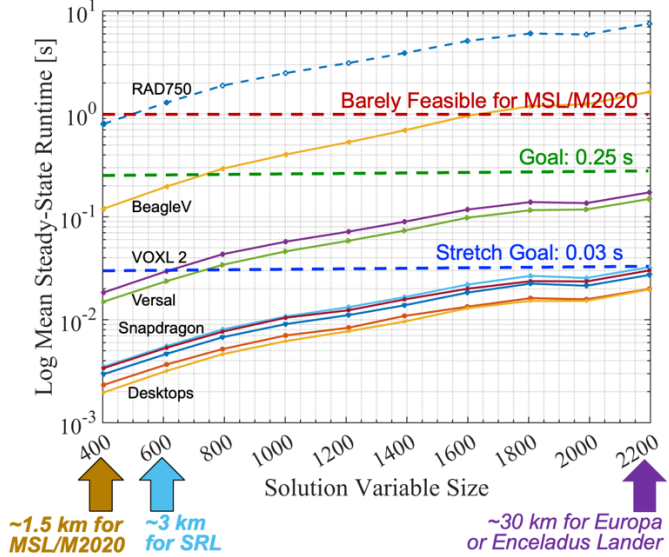

Fig. 1. *GFOLD Execution Performance on Legacy and Next-Generation Processors.* The log mean steady-state runtime of the GFOLD algorithm is shown as a function of solution variable size across various processors, including RAD750, BeagleV (RISC-V), VOXL2, Versal, Snapdragon, and desktop-class CPUs (Intel i7). The runtime thresholds for key mission profiles are marked: 0.25 s as the baseline goal for onboard real-time optimization, and 0.03 s as the stretch goal for ultra-low-latency guidance. RAD750 is barely feasible for MSL/M2020-scale diverts (~400 variables, ~1.5 km), while next-gen processors like Snapdragon achieve sub-0.03 s runtimes even for large-scale problems (~2200 variables, ~30 km) relevant to Europa or Enceladus landers. These results validate that modern multi-core architectures enable real-time, adaptive trajectory optimization for current and future planetary missions with increasing mission complexity and divert distances.

III. IDENTIFYING CRITICAL FAULT-PRONE STAGES IN GFOLD COMPUTATION

The GFOLD algorithm's convex optimization procedure comprises three primary computational stages: initialization, iterative gradient computations, and final constraint validation. To assess fault sensitivity across these stages, we performed controlled fault injection experiments involving single-bit flips at each stage and measured the resulting impact on algorithmic success.

Figure 2 presents the *Success Probability Distribution Across Trials* based on 100 Monte Carlo simulations per configuration. A bit flip introduced during the gradient computation stage causes the most severe degradation in performance—dropping success probability from 1.00 to 0.00 with a single fault. This sharp decline highlights the vulnerability of the gradient loop, where iterative floating-point operations amplify even minor numerical errors, potentially leading to divergence or invalid solutions.

In contrast, bit flips during the initialization stage cause a moderate reduction in robustness, reducing the success rate to 0.62. These errors typically introduce inaccuracies in the problem setup or solver parameters, which may degrade convergence but often still yield valid results. Faults introduced during the final constraint validation stage show minimal impact, with the success rate remaining at 0.88. However, these faults are particularly problematic — despite causing bit-level corruption, 88% of these errors result in outputs that appear numerically indistinguishable from correct solutions. This phenomenon represents a significant loss of fault observability, where traditional output-based fault detection or majority voting may fail to catch silent faults.

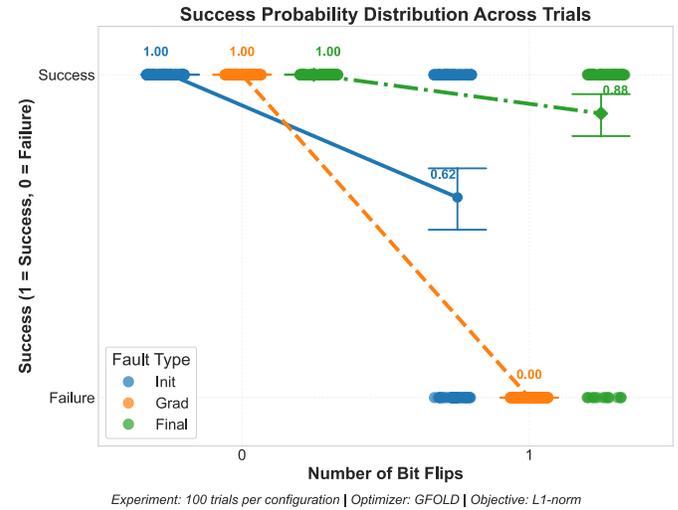

Fig. 2. *Success Probability Distribution Across Trials with Bit-Flip Fault Injection.* This figure shows the success rate (1 = success, 0 = failure) of the GFOLD optimizer under single-bit-flip faults introduced at three distinct computational stages: Initialization (blue), Gradient computation (orange), and Final constraint validation (green). Each data point represents the average over 100 Monte Carlo simulations, with error bars indicating variability. Gradient-stage faults result in a complete failure rate, while initialization faults reduce the success rate to 62%. Final-stage faults exhibit an 88% success rate, and often go undetected—highlighting reduced fault observability and the need for vector-based arbitration to improve reliability in mission-critical applications.

These insights have two major implications for fault-tolerant architecture design. First, protection mechanisms such as redundancy, selective checkpointing, or real-time error correction should be prioritized for the gradient computation stage, where undetected faults have the highest risk of mission failure. Second, the lack of observability in final-stage faults underscores the limitations of scalar output voting. To address this, we propose enhancing the multi-core voting mechanism by using vectorized intermediate outputs from all three GFOLD stages—initialization, gradient, and final validation. This approach would increase detection granularity and enable voting strategies that reflect internal consistency across solver iterations, rather than relying solely on final output comparisons.

Looking ahead, future missions will require adaptive guidance, where GFOLD runs multiple times during descent, incorporating updated state estimates from sensors and navigation data [1]. As the number of optimization runs increases, so does cumulative fault exposure—making robust, stage-aware protection strategies essential to ensuring reliable, real-time trajectory optimization in deep space operations.

## IV. FAULT-TOLERANT COMPUTING AND ARBITER MULTI-CORE VOTING MECHANISM

Radiation-induced faults, such as single-event upsets (SEUs), pose a persistent risk to onboard spacecraft computing [26], [27]. To address this, we introduce ARBITER (Asynchronous Redundant Behavior Inspection for Trusted Execution and Recovery)—a Multi-Core Voting mechanism designed to enhance fault tolerance through parallel execution and real-time output arbitration. ARBITER enables resilient operation of both static computations, such as optimization solvers (e.g., GFOLD), and dynamic control loops, such as Attitude Control Systems (ACS), by continuously verifying computation integrity across multiple cores.

The ARBITER framework executes the same algorithm across redundant cores, compares the resulting outputs, and isolates faulty behavior when discrepancies arise. For voting purposes, M+2 independent cores must be running identical software in parallel to catch M separate faults.

This approach supports two operational modes. In static mode, used for deterministic computations like trajectory optimization, redundant subroutine outputs (e.g., GFOLD solutions) are compared at defined checkpoints. In dynamic mode, used for real-time control systems, such as a Proportional-Derivative (PD) controller in an ACS, ARBITER continuously arbitrates outputs at each timestep, maintaining stable control responses even when faults occur mid-operation.

Figure 3 illustrates the architecture of ARBITER applied to a fault-tolerant PD controller. Each core independently processes its input data and computes control outputs. The voting mechanism evaluates the outputs in real time, issuing the final control command and generating a fault report if inconsistencies are detected. This structure supports modular scalability and adaptability across different mission-critical subsystems.

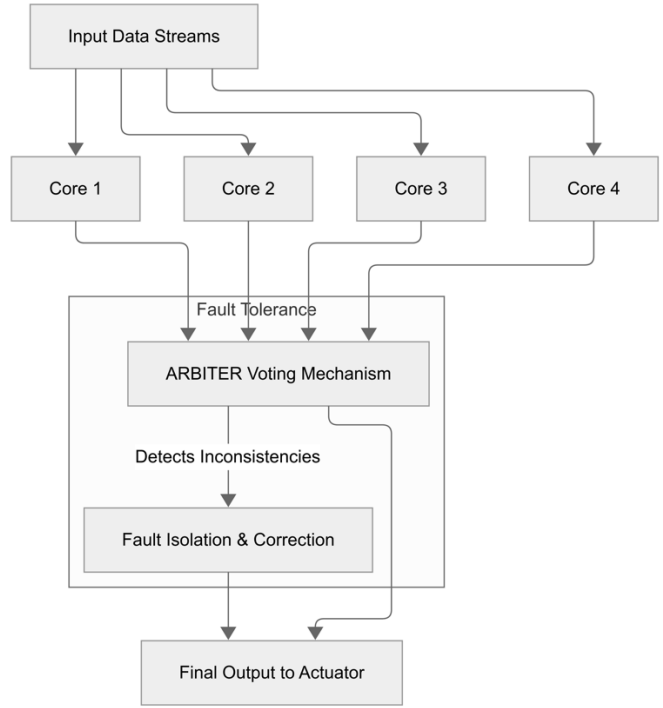

Fig. 3. *ARBITER Multi-Core Voting Architecture for Fault-Tolerant PD Control.* This diagram illustrates the ARBITER framework applied to a simplified closed-loop spacecraft control system. Input data streams are processed by four independent cores running identical PD control logic. Their outputs are fed to the ARBITER voting mechanism, which computes the final output when inconsistencies are detected. This design supports fault isolation and correction in radiation-prone environments, ensuring high-integrity control performance.

To evaluate ARBITER's fault resilience, we simulated a spacecraft with a double integrator dynamic model under a PD control law, executing at 8 Hz. Measurement noise was introduced on angle and rate feedback, and step commands were used to simulate typical maneuvering conditions. Bit-flip faults were injected on selected cores at specified time intervals to assess ARBITER's ability to isolate and correct corrupted outputs.

Figure 4 shows the closed-loop performance of the system under fault conditions. The top plot displays the arbitrated output, which tracks the ideal unfaulted trajectory with high fidelity despite injected faults. The bottom plot reveals the individual core outputs, highlighting divergence in core 1 after a fault at 15 seconds and in core 3 after 30 seconds. Despite these faults, ARBITER successfully rejects corrupted data and maintains a stable output, preserving system integrity throughout the test.

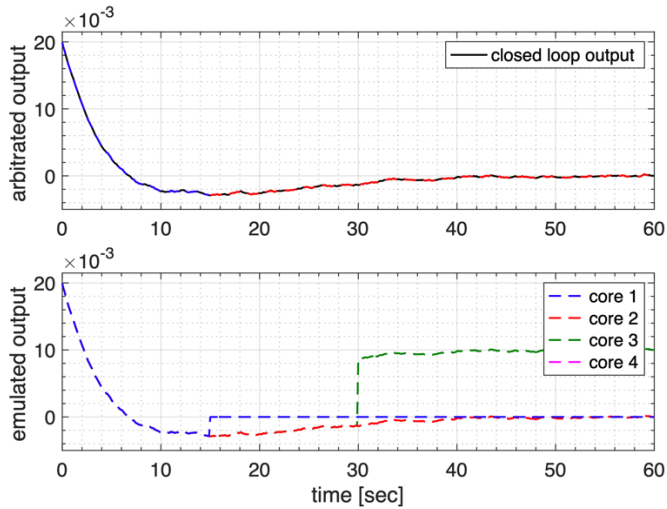

Fig. 4. *Dynamic Fault Tolerance Demonstration Using ARBITER.* **Top:** Arbitrated closed-loop output remains stable and converges toward zero despite injected faults. **Bottom:** Emulated outputs from individual cores show fault injections at 15 seconds (core 1) and 30 seconds (core 3), causing visible divergence. ARBITER isolates these faulty signals, preserving system performance. This test validates ARBITER's ability to maintain control integrity during real-time operations in the presence of transient faults.

When supporting closed-loop control, it is desired for the ARBITER to be called at the controller sampling rate and provide computation with sufficiently low latency so that the fault could be detected and arbitrated within the same sampling period. This avoids the ARBITER outputting an incorrect actuator command that could excite the spacecraft dynamics and create an unwanted transient event.

These results demonstrate ARBITER's effectiveness in detecting faults and preserving output correctness, which is especially critical for real-time autonomous operations in deep space environments. Future enhancements will include incorporating vector-based intermediate outputs for more granular fault detection.

The ARBITER voting mechanism can be hosted across several hardware configurations depending on the mission architecture. In the HPSC platform, ARBITER may be implemented directly within the multicore RISC-V clusters. In FPGA-based systems, such as LVS or GNC subsystems, ARBITER can be deployed in soft logic to minimize latency and maximize configurability. Alternatively, for missions requiring architectural isolation or tighter real-time guarantees, a dedicated radiation-hardened microcontroller can execute ARBITER alongside sensor fusion and actuator interface logic. The architecture is modular and can be adapted to meet power, fault-tolerance, and timing constraints of the flight processor environment.

## V. CONCLUSIONS AND FUTURE WORK

This paper establishes a scalable, high-performance, and fault-tolerant space computing architecture designed to support the computational demands of future autonomous planetary exploration missions. By implementing core algorithms—Landing Vision System (LVS) and Guidance for Fuel-Optimal Large Diverts (GFOLD)—on next-generation processors including HPSC, Snapdragon VOXL2, and AMD Xilinx Versal, we achieve substantial improvements in execution time, energy efficiency, and fault resilience. Experimental benchmarking demonstrates that GNC and LVS routines, previously constrained by the processing limits of RAD750 and Virtex-5 hardware, can now operate in real time with up to 250× speedup, meeting or exceeding future mission timing constraints.

We also introduce and validate ARBITER, a Multi-Core Voting (MV) fault mitigation system capable of detecting and resolving discrepancies through redundant execution. ARBITER is applied to both static scenarios (e.g., GFOLD subroutine execution) and dynamic closed-loop control scenarios (e.g., Attitude Control System). In addition, sensitivity analysis of GFOLD reveals that the gradient computation stage is particularly vulnerable to bit-flip faults, informing where targeted protection should be applied in fault-tolerant design. This fault observability analysis motivates further refinement of ARBITER through vector-based intermediate outputs, enabling higher resolution voting and real-time error diagnosis.

Our approach directly supports a wide range of future mission architectures. For high-gravity landers like Mars Sample Return and Ceres Lander, rapid trajectory re-computation and image-based hazard avoidance are required under stringent real-time constraints. For small-body missions, such as Comet Sample Return and Centaur Orbiter, gravity is lower so that operations can proceed at a lower rate and the real-time computational constraints are somewhat relaxed. However, here challenges arise from significant uncertainty (low-cost vehicles, complex sensing and state estimation, coarse maps, etc.) and slower-response controllers, so that onboard computing must accommodate flexible descent profiling and constraints, hazard avoidance, and risk-aware optimization. Furthermore, applications like in-space assembly, rendezvous, and formation flying require fault-resilient, uncertainty-aware decision-making embedded within constrained hardware. Our study has demonstrated encouraging results towards meeting these demands and enabling new levels of autonomy and mission assurance.

Future work would focus on several extensions: incorporating adaptive guidance via multiple G-FOLD executions during flight; enhancing ARBITER with vector outputs for finer-grained voting and fault detection; and extending optimization algorithms to handle non-convex and state-triggered constraints. These advancements will ensure that future

missions—from pinpoint landings on icy moons to precision orbital maneuvers in cislunar space—can be carried out with high autonomy, reliability, and performance.

ACKNOWLEDGMENT

The research was carried out at the Jet Propulsion Laboratory, California Institute of Technology, under a contract with the National Aeronautics and Space Administration (80NM0018D0004).

REFERENCES

[1] M. B. Quadrelli, L. J. Wood, J. E. Riedel, M. C. McHenry, M. Aung, L. A. Cangahuala, R. A. Volpe, P. M. Beauchamp, and J. A. Cutts, "Guidance, Navigation, and Control Technology Assessment for Future Planetary Science Missions," *Journal of Guidance, Control, and Dynamics*, vol. 38, no. 7, pp. 1165–1186, Jul. 2015.
[2] A. Banerjee, M. Mukherjee, S. Satpute, and G. Nikolakopoulos, "Resiliency in Space Autonomy: a Review," *Curr Robot Rep*, vol. 4, no. 1, pp. 1–12, Mar. 2023.
[3] L. D. Kennedy, "NASA Lunar Lander Reference Design," 2019.
[4] T. M. Lovelly, D. Bryan, K. Cheng, R. Kreynin, A. D. George, A. Gordon-Ross, and G. Mounce, "A framework to analyze processor architectures for next-generation on-board space computing," in *2014 IEEE Aerospace Conference*, 2014, pp. 1–10.
[5] J. Goodwill, C. Wilson, and J. MacKinnon, "Current AI technology in space," in *Precision Medicine for Long and Safe Permanence of Humans in Space*, Elsevier, 2025, pp. 239–250.
[6] G. Furano and A. Menicucci, "Roadmap for On-Board Processing and Data Handling Systems in Space," in *Dependable Multicore Architectures at Nanoscale*, M. Ottavi, D. Gizopoulos, and S. Pontarelli, Eds. Cham: Springer International Publishing, 2018, pp. 253–281.
[7] C. Yahnker, S. Ardito, J. Castillo-Rogez, A. Argueta, B. Morin, T. Canham, J. Butler, J. Gates, T. Pham, and A. Ferrer, "Vision and roadmap for the next generation of spaceflight computing," in *2024 IEEE Aerospace Conference*, 2024, pp. 1–9.
[8] J. Cong, J. Lau, G. Liu, S. Neuendorffer, P. Pan, K. Vissers, and Z. Zhang, "FPGA HLS Today: Successes, Challenges, and Opportunities," *ACM Trans. Reconfigurable Technol. Syst.*, vol. 15, no. 4, pp. 1–42, Dec. 2022.
[9] B. Schwaller, S. Holtzman, and A. D. George, "Emulation-based performance studies on the HPSC space processor," in *2019 IEEE Aerospace Conference*, 2019, pp. 1–11.
[10] A. Kalantari, A. Brinkman, K. Carpenter, M. Gildner, J. Jenkins, D. Newill-Smith, J. Seiden, A. Umali, and R. Mccormick, "Design, Prototype, and Performance Assessment of an Autonomous Manipulation System for Mars Sample Recovery Helicopter," in *2024 IEEE/RSJ International Conference on Intelligent Robots and Systems (IROS)*, 2024, pp. 2229–2236.
[11] W. Reid, T. Bartlett, A. Bouton, M. P. Strub, M. Newby, S. Gerdts, J. Martin, S. Moreland, and R. McCormick, "Planning and control for autonomous drives of the mars sample recovery helicopter," in *2024 IEEE Aerospace Conference*, 2024, pp. 1–11.
[12] N. Perryman, S. Sabogal, C. Wilson, and A. George, "Dependable DPU Architectures on AMD-Xilinx Versal Adaptive SoCs for Space Applications," *IEEE Transactions on Aerospace and Electronic Systems*, 2025.
[13] B. Gaide, D. Gaitonde, C. Ravishankar, and T. Bauer, "Xilinx Adaptive Compute Acceleration Platform: Versal™ Architecture," in *Proceedings of the 2019 ACM/SIGDA International Symposium on Field-Programmable Gate Arrays*, Seaside CA USA, 2019, pp. 84–93.
[14] W. A. Powell, "High-performance spaceflight computing (hpsc) project overview," in *Radiation Hardened Electronics Technology Conference (RHET) 2018*, 2018, no. GSFC-E-DAA-TN62651.
[15] L. Delvaux and L. Di Naro, "Autopilot and companion computer for unmanned aerial vehicle: Survey," 2023.
[16] N. Perryman, A. George, J. Goodwill, S. Sabogal, D. Wilson, and C. Wilson, "Comparative Analysis of Next-Generation Space Computing Applications on AMD-Xilinx Versal Architecture," *Journal of Aerospace Information Systems*, vol. 22, no. 2, pp. 103–115, Feb. 2025.
[17] M. Petry, G. Wuwer, A. Koch, P. Gest, M. Ghiglione, and M. Werner, "Accelerated deep-learning inference on the versal adaptive SoC in the space domain," in *2023 European Data Handling & Data Processing Conference (EDHPC)*, 2023, pp. 1–8.
[18] G. Lentaris, K. Maragos, I. Stratakos, L. Papadopoulos, O. Papanikolaou, D. Soudris, M. Lourakis, X. Zabulis, D. Gonzalez-Arjona, and G. Furano, "High-Performance Embedded Computing in Space: Evaluation of Platforms for Vision-Based Navigation," *Journal of Aerospace Information Systems*, vol. 15, no. 4, pp. 178–192, Apr. 2018.
[19] W. Powell, "NASA's vision for spaceflight computing," in *16th ESA Workshop on Avionics, Data, Control and Software Systems (ADCSS)*, 2022.
[20] A. E. Johnson, S. B. Aaron, H. Ansari, C. Bergh, H. Bourdu, J. Butler, J. Chang, R. Cheng, Y. Cheng, K. Clark, D. Clouse, R. Donnelly, K. Gostelow, W. Jay, M. Jordan, S. Mohan, J. Montgomery, J. Morrison, S. Schroeder, B. Shenker, G. Sun, N. Trawny, C. Umsted, G. Vaughan, M. Ravine, J. Schaffner, J. Shamah, and J. Zheng, "Mars 2020 Lander Vision System Flight Performance," in *AIAA SCITECH 2022 Forum*, San Diego, CA & Virtual, 2022.
[21] S. B. Aaron, Y. Cheng, N. Trawny, S. Mohan, J. Montgomery, H. Ansari, K. Smith, A. E. Johnson, J. Goguen, and J. Zheng, "Camera Simulation For Perseverance Rover's Lander Vision System," in *AIAA SCITECH 2022 Forum*, San Diego, CA & Virtual, 2022.
[22] B. Acikmese, M. Aung, J. Casoliva, S. Mohan, A. Johnson, D. Scharf, D. Masten, J. Scotkin, A. Wolf, and M. W. Regehr, "Flight testing of trajectories computed by G-FOLD: Fuel optimal large divert guidance algorithm for planetary landing," in *AAS/AIAA spaceflight mechanics meeting*, 2013, p. 386.
[23] D. P. Scharf, M. W. Regehr, G. M. Vaughan, J. Benito, H. Ansari, M. Aung, A. Johnson, J. Casoliva, S. Mohan, and D. Dueri, "ADAPT demonstrations of onboard large-divert guidance with a VTVL rocket," in *2014 IEEE aerospace conference*, 2014, pp. 1–18.
[24] D. P. Scharf, B. Açıkmeşe, D. Dueri, J. Benito, and J. Casoliva, "Implementation and Experimental Demonstration of Onboard Powered-Descent Guidance," *Journal of Guidance, Control, and Dynamics*, vol. 40, no. 2, pp. 213–229, Feb. 2017.
[25] B. Acikmese, J. Casoliva, J. M. Carson, and L. Blackmore, "G-fold: A real-time implementable fuel optimal large divert guidance algorithm for planetary pinpoint landing," *Concepts and Approaches for Mars Exploration*, vol. 1679, p. 4193, 2012.
[26] R. Ecoffet, "Overview of in-orbit radiation induced spacecraft anomalies," *IEEE Transactions on Nuclear Science*, vol. 60, no. 3, pp. 1791–1815, 2013.
[27] S. M. Foulds, "Study of single event upsets (SEUS) a survey and analysis," PhD Thesis, Toronto Metropolitan University, 2013.